\documentclass[letterpaper, 10 pt, conference]{ieeeconf}  

\IEEEoverridecommandlockouts                              

\usepackage{graphics} 
\usepackage{epsfig} 
\usepackage{mathptmx} 
\usepackage{times} 
\usepackage{amsmath} 
\usepackage{amssymb}  
\usepackage{graphicx}
\usepackage{makecell}
\usepackage{url}
\usepackage{cite}
\usepackage{threeparttable}
\usepackage{hyperref}
\hypersetup{hidelinks,
	colorlinks=true,
	allcolors=blue,
	breaklinks=true}
\usepackage[switch]{lineno}

\title{\LARGE \bf

MSSP : A Versatile Multi-Scenario Adaptable Intelligent Robot Simulation Platform Based on LIDAR-Inertial Fusion
}

\author{Qiyan Li, Chang Wu$^{*}$, Yifei Yuan, Yuan You 
\thanks{To protect intellectual property, the source code of the simulation platform will be made publicly available following the acceptance of the paper.}
\thanks{Qiyan Li, Chang Wu, Yifei Yuan are with the Institute of Information and Communication Engineering, University of Electronic Science and Technology of China (UESTC), Chengdu 611731, China (e-mail: qyan.lieestd@gmail.com, changwu@uestc.edu.cn; yuanyf1998@126.com)(Corresponding author: Chang Wu.).
        }%
\thanks{Yuan You is with the XGRIDS, Shenzhen 518051, China (e-mail: yuan.you@xgrids.com).
        }%
}

\usepackage{amssymb}
\usepackage{pifont}

\providecommand{\keywords}[1]{\textbf{\textit{Index terms---}} #1}

\makeatletter

\newcommand{\Rmnum}[1]{\expandafter\@slowromancap\romannumeral #1@}
\makeatother

\begin{document}

\maketitle
\thispagestyle{empty}
\pagestyle{empty}

\begin{abstract}

This letter presents a multi-scenario adaptable intelligent robot simulation platform based on  LIDAR-inertial fusion, with three main features: (1 The platform includes an versatile robot model that can be freely controlled through manual control or autonomous tracking. This model is equipped with various types of LIDAR and Inertial Measurement Unit (IMU), providing ground truth information with absolute accuracy. (2 The platform provides a collection of simulation environments with diverse characteristic information and supports developers in customizing and modifying environments according to their needs. (3 The platform supports evaluation of localization performance for SLAM frameworks. Ground truth with absolute accuracy eliminates the inherent errors of global positioning sensors present in real experiments, facilitating detailed analysis and evaluation of the algorithms. By utilizing the simulation platform, developers can overcome the limitations of real environments and datasets, enabling fine-grained analysis and evaluation of mainstream SLAM algorithms in various environments. Experiments conducted in different environments and with different LIDARs demonstrate the wide applicability and practicality of our simulation platform. The implementation of the simulation platform is open-sourced on Github.

\end{abstract}


\section{INTRODUCTION}

Intelligent robots equipped with sensor suites find wide applications across various domains such as service guidance\cite{service_guidance}, field exploration\cite{field_exploration}, agricultural supervision\cite{agriculture}, and factory logistics\cite{factory_logistic}. Due to the high-precision ranging capabilities and adaptability to diverse environments\cite{precision_LIDAR}, LIDAR SLAM technology has rapidly developed and become the mainstream solution for complex tasks. Despite advancements in LIDAR SLAM field and the emergence of high-performance frameworks, several challenges persist : (1 The emergence of various LIDARs including mechanical rotating LIDARs and recent solid-state LIDARs makes it impractical to acquire all types of sensors for validation and testing, due to the cost and limitations in hardware procurement and deployment. This has resulted in relatively insufficient research on the adaptability of existing algorithms to different sensors, thereby restricting in-depth exploration of algorithm performance and robustness. (2 As application scenarios for intelligent robots become increasingly complex and varied, traditional data collection methods are repetitive and cumbersome, failing to cover all potential scenarios. This severely limits the development and widespread application of intelligent robot technology, such as jungle and cave exploration\cite{jungle_exploration,cave_exploration}, as well as deep-sea and deep-space exploration\cite{deep_sea,deep_space}. (3 Current validation and evaluation of SLAM algorithms rely primarily on ground truth provided by GPS or other global positioning devices\cite{lio-sam,fast-lio2}. This approach not only incurs hardware deployment and device costs but also introduces inherent biases due to errors in the global positioning sensors themselves. Consequently, the credibility of experimental results is affected, limiting detailed studies of algorithm performance.
	
However, to address the current challenges in LIDAR SLAM, simulation technology offers an efficient and advantageous solution. By using precise simulation models, we can emulate various LIDAR sensors in virtual environments, eliminating the costs and hardware deployment issues associated with physical devices. Moreover, advanced simulation technology allows us to create customized virtual environments as needed, avoiding the time and cost constraints of data collection and the difficulties or impossibilities of collecting data in extreme scenarios\cite{exterme_envir,ocean_dataset}. Additionally, simulation environments can provide ground truth information with absolute accuracy, enhancing the credibility of SLAM algorithm accuracy evaluations and enabling more detailed assessments of algorithm performance. Therefore, in this paper, we propose a multi-scenario adaptable intelligent robot simulation platform based on LIDAR-inertial fusion, named MSSP. Specifically, the innovations of our platform include:

\begin{enumerate}
    \item We have designed and built a multi-scenario adaptable intelligent robot simulation platform based on LIDAR-inertial fusion in Gazebo, called MSSP. This platform is equipped with mechanical and solid-state LIDARs and IMU sensors, supporting both manual control and autonomous tracking modes. Additionally, it provides ground truth information with absolute accuracy of the robot's position in real-time through Gazebo plugins. The platform is highly user-friendly and extremely extensible.

    \item The simulation platform supports the convenient import and custom creation of various types of virtual simulation environments, including various types of indoor and outdoor environments, as well as special scenarios such as degraded, unstructured, and dynamic environments. This capability overcomes the limitations of datasets and real-world environments, covering mainstream research and application scenarios for intelligent robots.

    \item The simulation platform supports the validation and evaluation of LIDAR SLAM algorithms. By utilizing the ground truth information with absolute accuracy provided by simulation, we could perform detailed analyses of the algorithms, avoiding the inherent errors present in global positioning devices. Our analysis and evaluation of various SLAM algorithms across different scenarios demonstrate the platform's effectiveness and practical value.

    \item We make the simulation platform publicly available on GitHub to share our findings and contribute to the open-source community. In the future, developers can expand the platform's functionalities according to their specific needs, such as adding new sensors or virtual environments, thereby simplifying the process of algorithm development and validation.
\end{enumerate}

\section{SYSTEM OVERVIEW}

\begin{figure*}
    \centering
    \includegraphics[width=18cm]{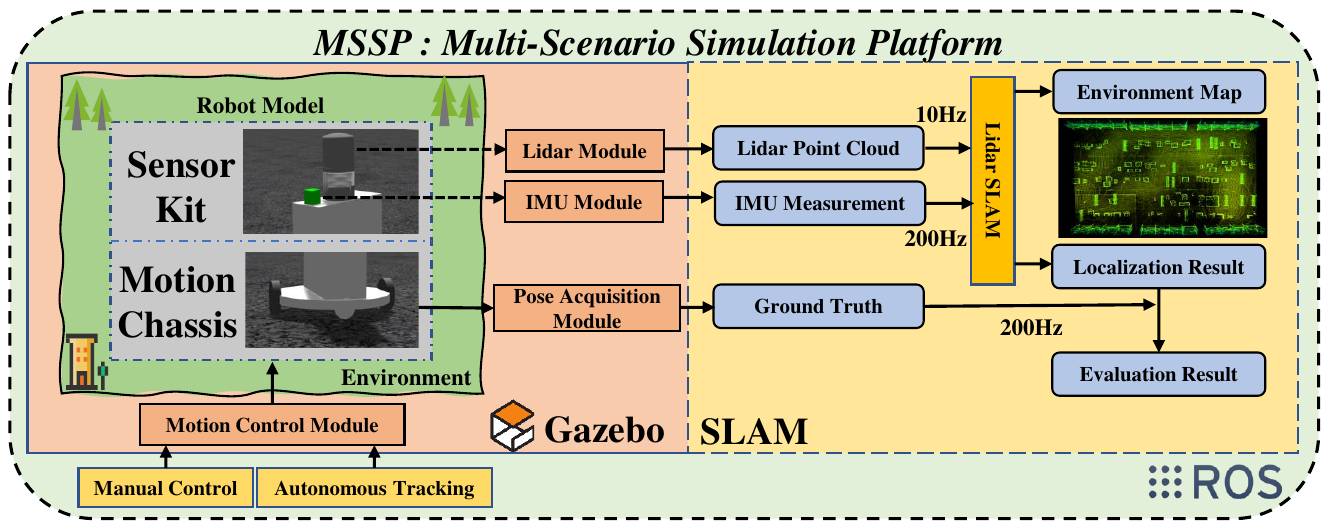}
    \caption{System architecture of MSSP simulation platform. This platform comprises two
core components : the robot system simulation(shown in left pink rectangular box) and the SLAM algorithm evaluation(shown in right yellow rectangular box)}
    \label{system overview}
\end{figure*}

Figure \ref{system overview} clearly illustrates the system architecture of MSSP simulation platform, which primarily comprises two core components: the robot system simulation and the SLAM algorithm evaluation, both implemented within the ROS environment\cite{ros}.

The robot system simulation comprises two key modules: the robot model and the simulation environment, both implemented in Gazebo. Leveraging Gazebo simulation tool, the robot model and simulation environment are successfully loaded and can be customized, edited, and modified through a graphical user interface (GUI) or programming. The Gazebo simulation environment is constructed using various predefined models arranged in specific layouts to simulate real-world scenarios. The robot model consists of a motion chassis and a sensor suite: The motion chassis, controlled through a motion control module (implemented via Gazebo's built-in model differential control plugin), receives external inputs, allowing users to control the robot's movement within the simulation environment via manual control or autonomous tracking. During the robot's movement, the onboard sensor suite (including LIDAR and IMU) generates virtual sensor data within Gazebo through LIDAR and IMU modules (implemented via Gazebo's sensor plugins) to simulate real-world perception capabilities, achieving detailed scans of the external environment. The generated sensor data, primarily comprising point clouds and IMU measurements, are transmitted to subsequent LIDAR SLAM algorithm processing modules to obtain environmental map and robot localization information. Then, leveraging the system's pose acquisition module (implemented via Gazebo's built-in model plugin), ground truth information with absolute accuracy is obtained. Based on this, we can accurately and comprehensively evaluate the performance of various SLAM algorithms in different environments, providing a solid experimental foundation for further research and practical applications.

In the following sections, we will elaborate on the specific details of the robot model creation and robot motion control methods. Details regarding the configuration of the simulation environment and testing of SLAM algorithms will be covered in Section \Rmnum{3} and \Rmnum{4}, respectively. 

\subsection{Robot Model Creation}

\begin{figure}
    \centering
    \includegraphics[width=8cm]{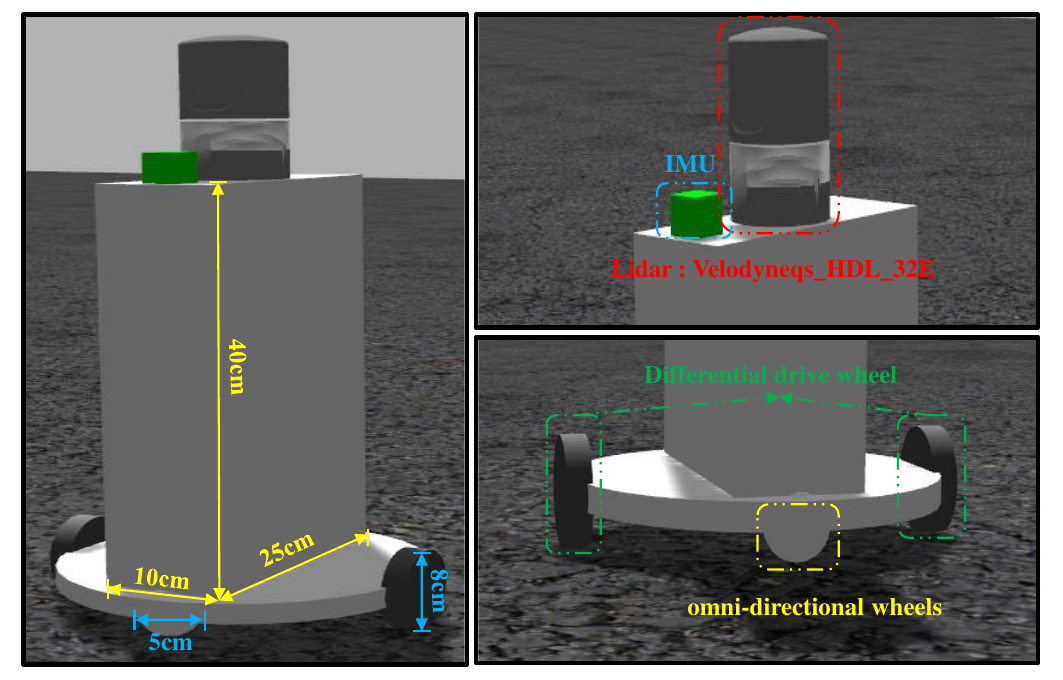}
    \caption{System architecture of the robot simulation model. The left image shows the overall structure and detailed dimensions of the robot model, while the right one presents detailed information on the sensor suite and motion chassis.}
    \label{robot model}
\end{figure}

The simulation robot model is constructed using .xacro files, which are organized in XML format. These files allow for adjusting the specifications and configurations of the robot through code editing and parameter settings, thereby simplifying the management of complex robot models. Figure \ref {robot model} illustrates the detailed system architecture of the simulated robot model, comprising two core components: the motion chassis and the sensor suite. The design of the motion chassis is inherited from Diff-Robot (\href{https://www.mahaofei.com/post/fc92db80}{diff\_wheeled\_robot}), featuring two actively differential driven wheels and two passive omni-directional wheels, with radii of 4cm and 2.5cm, respectively. The differential control of the drive wheels enables complex driving and steering operations, while the presence of omni-directional wheels ensures the stability of the robot, enabling it to achieve stable and reliable motion in the Gazebo environment.

Furthermore, to guarantee that the sensors can possess an optimal sensing range and can perform the most effective environmental scanning, the sensor suite, comprising LIDAR and IMU sensors, is mounted on a platform with dimensions of 40 cm × 10 cm × 25 cm. This platform is rigidly connected to the motion chassis.

\begin{enumerate}
    \item \textbf{LIDAR}: The robot model is equipped with various types of LIDAR, including traditional mechanical rotating LIDARs(e.g., Velodyne\_HDL\_32E (\href{https://bitbucket.org/DataspeedInc/velodyne\_simulator}{velodyne\_simulator})), and emerging solid-state LIDARs (e.g., Livox series (\href{https://github.com/Livox-SDK/livox\_laser\_simulation}{livox\_laser\_simulation})). Users can select an appropriate LIDAR based on their specific requirements. They are positioned at different locations on the platform(with the mechanical LIDAR placed in the center and the solid-state LIDAR positioned at the front) to achieve maximum environmental coverage, generating environmental scan data at 10 Hz.

    \item \textbf{IMU}: The robot model is equipped with a 9-axis IMU sensor providing system acceleration, angular velocity, and magnetic field direction information. During simulation, the frequency of IMU can be adjusted as needed and is set to 200 Hz in our system.
\end{enumerate}

In simulation platform, LIDAR and IMU sensor data have been augmented with appropriate noise to simulate real-world conditions. The noise can be adjusted or removed according to experimental requirements. In contrast to practical testing and validation of SLAM algorithms, the simulation platform obtains ground truth using gazebo built-in model plugin(libgazebo\_ros\_p3d.so), instead of simulated data from GPS or other global positioning devices. This plugin computes the position of the robot's body frame relative to a fixed reference frame in the simulated world. It publishes the robot's three-dimensional coordinates (position information) and quaternion orientation information through the ROS interface in standard message formats at a fixed frequency (set to maximum output frequency of all sensors in our platform, 200Hz). This approach ensures that the ground truth is absolutely accurate and error-free.

\subsection{Robot Motion Control}

Effective motion control is crucial for the successful simulation of intelligent robots. In this letter, we adopt the differential drive model\cite{differential_robot_model} as the robot motion control mechanism, designing and implementing both manual control and autonomous tracking modes to generate control signals, ensuring the robot moves precisely according to user intentions. Differential drive control is a widely used control strategy in wheeled robots, based on the principle of independently adjusting the speeds of the robot's wheels on either side. By manually controlling the rotational speed of different wheels, the robot's path and direction can be altered. The manual control mode relies on proactive human input, providing a simple and flexible way to achieve basic robot motion and control, suitable for various experimental scenarios. On the other hand, autonomous tracking control sacrifices some flexibility but enables more precise and smooth robot control and path navigation, optimizing the robot's performance in complex environments. In the following sections, we will elaborate on the principles and specific implementations of these two motion control methods.

\subsubsection{Manual Control of the Robot}
    
The manual control of the robot is achieved by defining a mapping relationship between keyboard keys and robot actions, resulting in an intuitive control interface. This allows the operator to adjust the robot's direction and speed through simple keyboard operations. Subsequently, velocity and orientation information is published in the Twist message format, which is received by the robot's motion chassis and converted into specific wheel speed control signals. This ensures that the robot can move precisely according to the operator's intentions.

\subsubsection{Autonomous Tracking of the Robot}
    
Autonomous tracking of the robot combines the Bspline path generation algorithm\cite{Bspline} and the Pure Pursuit tracking algorithm\cite{PP_track} to enhance control accuracy. In the path generation module, using the model insertion function of the Gazebo simulation tool, key discrete points on the path are first visually marked, which defines the shape of the curve. Subsequently, these discrete points are used to generate a two-dimensional cubic Bspline curve to ensure smoothness and continuity of the path, with 5000 points sampled for subsequent use in the autonomous tracking module. In the tracking module, the system obtains the robot's current position and orientation information by subscribing to odometry messages in real-time. This information, combined with the target path points in the generated trajectory, allows the system to calculate the angular difference and distance between the current position and the target point. Based on the system's predefined linear velocity (set to 0.2 or 0.4 m/s as needed in our platform), the system computes the angular velocity required for directional control and publishes these control signals to the robot's motion control chassis to guide the robot towards the target path points. When the robot's distance to the target path point falls below a threshold(set to 0.01 meters in the platform, adjustable based on practical needs), the system updates the target point and continues the tracking operation. This precise control mechanism ensures that the robot can follow the predetermined B-spline curve accurately.

\begin{figure}
\centering
\includegraphics[width=8cm]{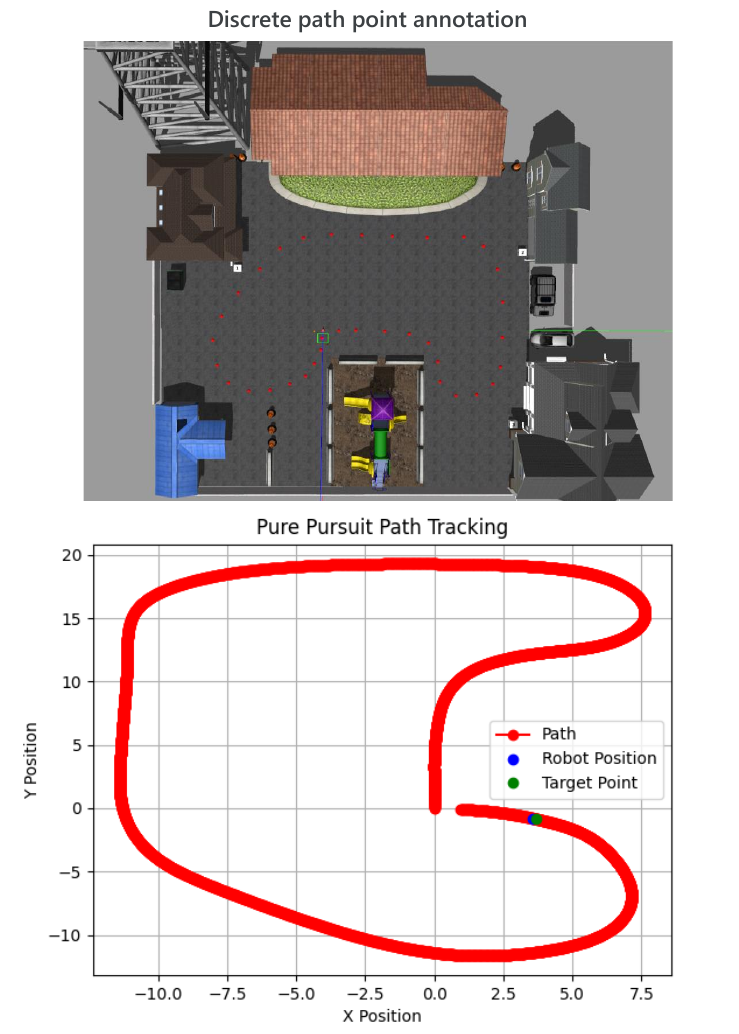}
\caption{Bspline Path Generation and Pure Pursuit Algorithm for Robot Autonomous Tracking}
\label{BPP}
\end{figure}

Figure \ref{BPP} illustrates the detailed process of autonomous tracking. In the above image, the red points are the user-defined discrete path points, and the yellow point indicates the end of the path. In the following image, the red curve represents the generated target path, and the pure pursuit tracking algorithm achieves autonomous tracking by continuously controlling the robot (Robot Position) to approach the target point (Target Point) and updating the target point persistently.

\section{Simulation Environment}

The generation and import of the simulation environment in Gazebo are achieved through .world files. These files are written in XML format and define various elements in the simulation world, including collections of robots and objects (such as buildings, tables, chairs, and trees), as well as global parameter information (such as sky, lighting, and physical properties). 
Users can customize the creation of simulation environments according to their specific needs or directly import publicly available environments provided by the official Gazebo repository or the open-source community to facilitate convenient development.

Users can flexibly create or configure the desired simulation environment through a graphical user interface (GUI) or programming. Configuring and creating virtual environments through Gazebo's graphical user interface (GUI) is convenient and intuitive. Using this approach, developers can visually select and arrange predefined models from Gazebo's model library, and customize scenes by combining these models in specific ways. Gazebo's built-in model library covers a wide range of predefined models, including environmental elements like buildings, trees, roads, and bridges. These predefined model files are organized in the Simulation Description Format (SDF), allowing developers to precisely define various aspects of the models or create specific models by writing or modifying SDF files according to their needs. For highly customized models, such as complex environmental structures, developers can create the 3D models using software like Blender or Maya, and then import them into Gazebo for further development.

Alternatively, directly editing .world files provides a higher level of customization. In .world files, developers can manually add or modify environmental features, and meticulously configure attributes of simulation objects such as position, size, material, and dynamic behaviors. Both methods empower developers to customize or generate virtual environments according to their specific requirements, freeing them from the constraints of datasets and real-world scenes. 

Gazebo provides a variety of official simulation world examples and comes equipped with numerous predefined model files, facilitating developers to configure simulation environments according to their preferences. Additionally, the open-source community has contributed a wealth of simulated world environments. We select 10 environments with distinct features suitable for algorithm validation as scenarios for subsequent SLAM algorithm evaluation. Due to variations in performance of different SLAM algorithms in indoor and outdoor environments, we categorize the candidate scenarios into indoor and outdoor types to comprehensively assess algorithm performance across different scene types.

\begin{figure*}
\centering
\includegraphics[width=18.5cm]{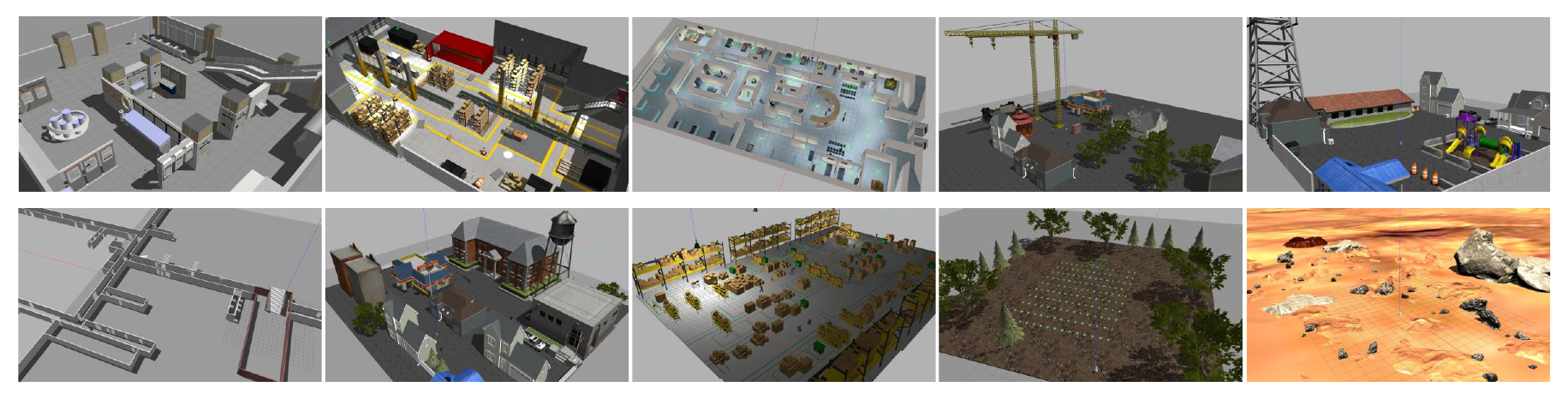}
\caption{Specific Schematic Diagrams of 10 Simulated Environments, from top left to bottom right, respectively corresponding to museum, factory, hospital, neighborhood, courtyard, corridor, district, warehouse, farmland, and desert scene.}
\label{environment_pic}
\end{figure*}

Furthermore, localization and mapping in non-ideal scenarios pose significant challenges in LIDAR SLAM research, such as occlusion in dynamic scenes, feature scarcity in unstructured environments, and algorithm degradation in indoor long corridors. To support algorithm validation under such conditions, the aforementioned 10 environments includes 5 non-ideal virtual simulation environments, including degraded scenarios, dynamic scenes, and unstructured environments. Detailed information regarding the type, and feature descriptions of each scene is provided in Table \ref{environment}, and the specific schematic diagrams of each scene are depicted in Figure \ref{environment_pic}.

\begin{table*}[]
\caption{Detailed information of 10 Virtual Simulated Environments}
\begin{tabular}{c|ccccc}
\hline
\textbf{Index} & \textbf{Environment} & \textbf{Type} & \textbf{Scale} & \textbf{Features} & \textbf{Description}                                                                      \\ \hline
\textbf{1}     & Museum\cite{museum}               & Indoor        & Smal           & Structural        & An indoor museum scene with stairs, exhibits, round tables, etc.            \\
\textbf{2}     & Factory\cite{factory-hospital}              & Indoor        & Medium         & Structural        & An indoor factory scene with stairs, conveyor belts, cargo, etc.                       \\
\textbf{3}     & Hospital\cite{factory-hospital}             & Indoor        & Large          & Structural        & An indoor hospital scene with several rooms, personnel, equipments, etc.    \\
\textbf{4}     & Corridor\cite{corridor}             & Indoor        & Large          & Degenerated       & A degenerated indoor scence with long straight corridors and spacious rooms, etc                       \\
\textbf{5}     & Warehouse\cite{warehouse}            & Indoor        & Large          & Dyanmic           & An indoor warehouse scene with dynamic workers, static goods, etc.                                  \\
\textbf{6}     & Neighborhood\cite{neighborhood, courtyard}         & Outdoor       & Medium         & Structural        & An outdoor neighborhood scene with several houses, trees, high-rise tower cranes, etc.               \\
\textbf{7}     & Courtyard\cite{neighborhood,courtyard}            & Outdoor       & Medium         & Structural        & An outdoor courtyard scene with houses, cars, high-rise signal towers, etc. \\
\textbf{8}     & District\cite{District}             & Outdoor       & Meidum         & Dyanmic           & An outdoor district scene with dynamic pedestrians, static houses, etc.                            \\
\textbf{9}     & Farmland\cite{farmland}             & Outdoor       & Small          & Unstructured      & A sparsely featured farmland scene with several trees and low-lying crops, etc.                 \\
\textbf{10}    & Desert\cite{mars}               & Outdoor       & Medium         & Unstructured      & A sparsely featured desert scene with undulating terrain and scattered rocks, etc.                      \\ \hline
\end{tabular}
\label{environment}
\end{table*}

\section{Experiment}

We conduct extensive validation experiments on mainstream LIDAR SLAM methods using the simulated platform we designed. The algorithm simulation and evaluation are conducted within the ROS system. We implemented our simulation platform and ran algorithms on a desktop with an Intel i7-13700F CPU (16 cores, 4.1GHz), 32GB RAM, and an NVIDIA GeForce RTX 3060 GPU. Due to significant performance variations in algorithms resulting from different LIDAR scanning patterns, we design experiments in two groups: mechanical LIDAR experiments and solid-state LIDAR experiments. Multiple sequences are selected and recorded from section \Rmnum{3}'s world environments, encompassing various types such as indoor and outdoor, structured and challenging, mechanical and solid-state modalities, and multiple control modes including manual controlling and autonomous tracking. Each sequence includes LIDAR and IMU sensor data, as well as ground truth information provided by Gazebo plugins. Basic information for each sequence can be found in the experimental results tables (Table \ref{mechanical table}, Table \ref{solid-state-table}).

\subsection{Experiments on Mechanical LIDARs}

Table \ref{mechanical table} presents detailed information about the test paths and evaluation results on A-LOAM\cite{aloam}, Lego-LOAM\cite{lego-loam}, LIO-SAM\cite{lio-sam}, FAST-LIO2\cite{fast-lio2}, Faster-LIO\cite{faster-lio}, Voxel-Map\cite{voxel-map}, and Point-LIO\cite{point-lio} algorithms. These paths encompass various environmental features used for testing SLAM algorithms, including indoor and outdoor, structured, dynamic, and degraded scenarios. These algorithms and scenarios are sufficient for demonstrating the extensive applicability and practical value of our platform. In our experiments, we use absolute pose error (APE) to quantify localization accuracy. The calculation of localization error was based on the EVO tool, an open-source trajectory error evaluation toolbox\cite{evo}.

\begin{table*}[]
\caption{Comparison of localization accuracy of mechanical LIDAR experiments (APE in meters)}
\begin{threeparttable}
\begin{tabular}{ccccccccccc}
\hline
\multicolumn{4}{c|}{\textbf{Sequence Description}}                                                             & \multicolumn{7}{c}{\textbf{Mainstream SLAM Algorithms Localization Error (Absolute Pose Error, APE)}}                                        \\ \hline
\textbf{Index}                  & \textbf{Sequence} & \textbf{Control} & \multicolumn{1}{c|}{\textbf{Feature}} & \textbf{A-LOAM} & \textbf{Lego-LOAM} & \textbf{LIO-SAM} & \textbf{Fast-LIO2} & \textbf{Faster-LIO} & \textbf{Voxel-Map} & \textbf{Point-LIO} \\ \hline
\multicolumn{1}{c|}{\textbf{1}} & Neighborhood      & Auto             & Structural                            & 0.027075        & 0.049364           & 0.018253         & 0.019316           & \textbf{0.014856}   & 0.046942           & 0.082936           \\
\multicolumn{1}{c|}{\textbf{2}} & Factory           & Manual           & Structural                            & 0.133986        & 0.138376           & 0.032304         & 0.030365           & \textbf{0.018654}   & 0.082183           & 0.061976           \\
\multicolumn{1}{c|}{\textbf{3}} & Hospital          & Auto             & Structural                            & 0.584886        & 1.459633           & 0.076913         & 0.067032           & 0.082596            & \textbf{0.059928}  & 0.249101           \\
\multicolumn{1}{c|}{\textbf{4}} & Courtyard         & Auto             & Structural                            & 0.022022        & 0.046197           & 0.018542         & 0.013465           & \textbf{0.011208}   & 0.033209           & 0.048859           \\
\multicolumn{1}{c|}{\textbf{5}} & Corridor          & Auto             & Structural                            & 0.165732        & 2.222618           & 0.244725         & 0.097725           & \textbf{0.095378}   & 0.096622           & 0.144621           \\
\multicolumn{1}{c|}{\textbf{6}} & Farmland          & Auto             & Structural                            & 0.062674        & 1.439014           & 0.023416         & \textbf{0.017185}  & 0.019390            & 0.059846           & 1.695804           \\
\multicolumn{1}{c|}{\textbf{7}} & District          & Auto             & Dynamic                               & 0.113169        & 0.045074           & 0.050324         & \textbf{0.044498}  & 0.072511            & 0.060674           & 0.064063           \\
\multicolumn{1}{c|}{\textbf{8}} & Warehouse         & Manual           & Dynamic                               & 0.050993        & 0.057560           & 0.044973         & 0.037286           & \textbf{0.022438}   & 0.044075           & 0.102127           \\ \hline
\end{tabular}
\label{mechanical table}
\begin{tablenotes}
        \footnotesize
        \item The bold values represent the algorithms with lowest positioning error in the current sequence.
      \end{tablenotes}
\end{threeparttable}
\end{table*}

As shown in Table \ref{mechanical table}, overall performance indicates that Fast-LIO2, Faster-LIO, and Voxel-Map perform notably well. Fast-LIO2 builds upon Fast-LIO\cite{fast-lio} by leveraging an incremental kd-Tree structure to efficiently represent large-scale, dense point cloud maps. This approach allows Fast-LIO2 to directly register raw LIDAR points into the map without the need for manually designed feature extraction modules. It can utilize more detailed environmental features, thereby enhancing the algorithm's localization and mapping performance. Faster-LIO introduces a sparse incremental voxel data structure instead of a tree structure to organize point clouds. This change significantly improves SLAM computational efficiency while maintaining system accuracy. In contrast to traditional point cloud maps, Voxel-Map proposes a precise and adaptive representation method for probabilistic voxel maps. By modeling uncertainty in plane features due to laser point measurements and pose estimation, Voxel-Map achieves accurate registration of LIDAR frames, thereby enhancing SLAM algorithm localization performance.

\begin{figure}
\centering
\includegraphics[width=8.5cm]{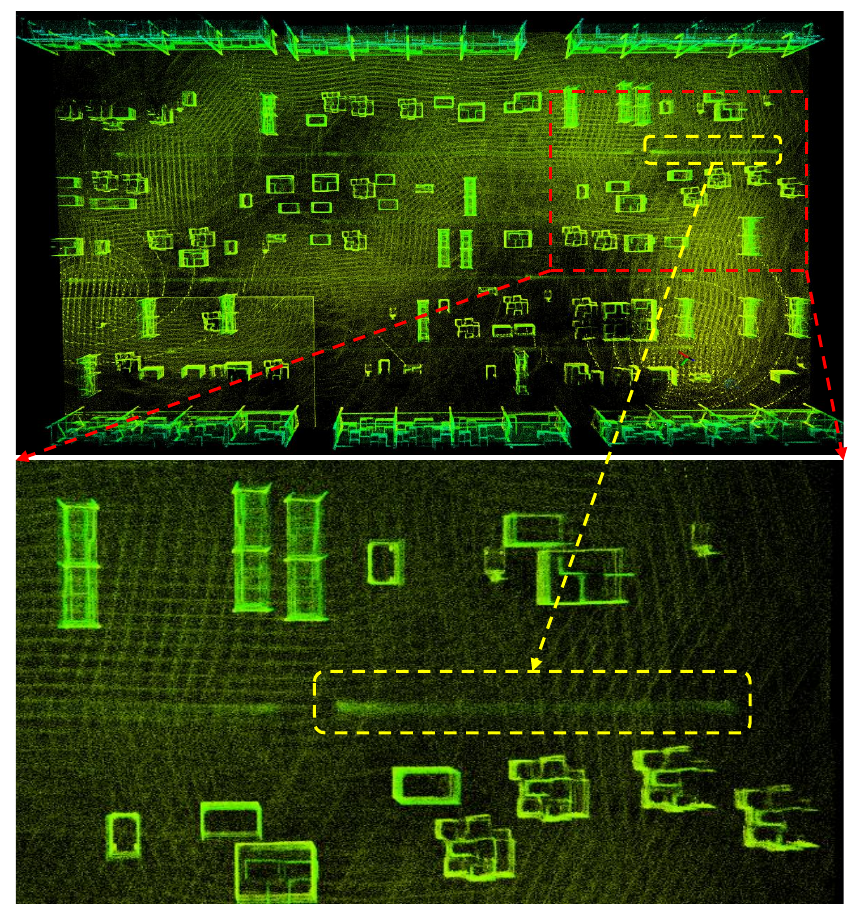}
\caption{Map constructed by Faster-LIO algorithm and "ghost" effects caused by moving personnel in warehouse sequence}
\label{dynamic}
\end{figure}

It's noteworthy that the sequences Warehouse and District depict dynamic scenes, each containing a small number of moving pedestrians or personnel. Mainstream SLAM systems and point cloud registration methods typically assume static environments\cite{rf-lio}. However, the presence of dynamic objects and significant occlusions can lead to feature tracking loss and consequently degrade pose estimation accuracy\cite{dynamic}. Moreover, accumulated scan data in point cloud maps may inadvertently capture unwanted paths due to dynamic objects, leading to “ghosting effects”. These "ghosts" are treated as obstacles in the map, severely affecting subsequent localization and navigation performance of intelligent robots\cite{ERASOR}. Figure \ref{dynamic} illustrates maps constructed by the Faster-LIO algorithm, highlighting "ghosting" effects caused by moving personnel (indicated by yellow dashed boxes). Given the rich environmental information and minimal occlusion effects from dynamic objects in sequences Warehouse-large and District-medium, their impact on SLAM algorithm localization accuracy is not significant.

\subsection{Experiments on Solid-state LIDARs}

Compared to mechanical LIDARs, solid-state LIDAR offers advantages in cost and portability. However, characteristics such as limited field of view and non-repetitive scanning introduce challenges that can degrade LIDAR SLAM algorithms based on solid-state LIDARs and make them more susceptible to occlusion from dynamic objects\cite{loam-livox}. We design a series of experiments to evaluate the localization and mapping performance of mainstream SLAM algorithms that support solid-state LIDARs in various environments. Our simulation platform offers seven models of LIDAR from Livox series for users to choose from, including Hap, Avia, Horizon, Mid 40, Mid 70, Mid 360 and Tele. Considering the product maturity and compatibility with publicly available algorithms, three LIDAR models with different parameters and performance characteristics from the Livox series are selected for algorithm performance evaluation. Table \ref{livox_series} provides detailed parameters information of the LIDARs used in Solid-state LIDAR experiments.

\begin{table}[]
\caption{Detailed Parameters of Solid-State LIDARs}
\begin{threeparttable}
\begin{tabular}{c|ccccc}
\hline
\textbf{LIDAR}   & \textbf{FoV\_H} & \textbf{FoV\_V} & \textbf{PointRate(point/s)} & \textbf{Range(m)} & \textbf{IMU}              \\ \hline
\textbf{Avia}   & 70.4°           & 77.2°           & 240,000            & 450              & \checkmark \\
\textbf{Hap}     & 120°            & 25°             & 452,000            & 150              & \ding{55}    \\
\textbf{Horizon} & 81.7°           & 25.1°           & 240,000            & 260              & \checkmark \\
\hline
\end{tabular}
\label{livox_series}
\begin{tablenotes}
        \footnotesize
        \item FoV\_H and FoV\_V represent the horizontal and vertical fields of view(FoV), respectively.
      \end{tablenotes}
\end{threeparttable}
\end{table}

\begin{figure*}
\centering
\includegraphics[width=18cm]{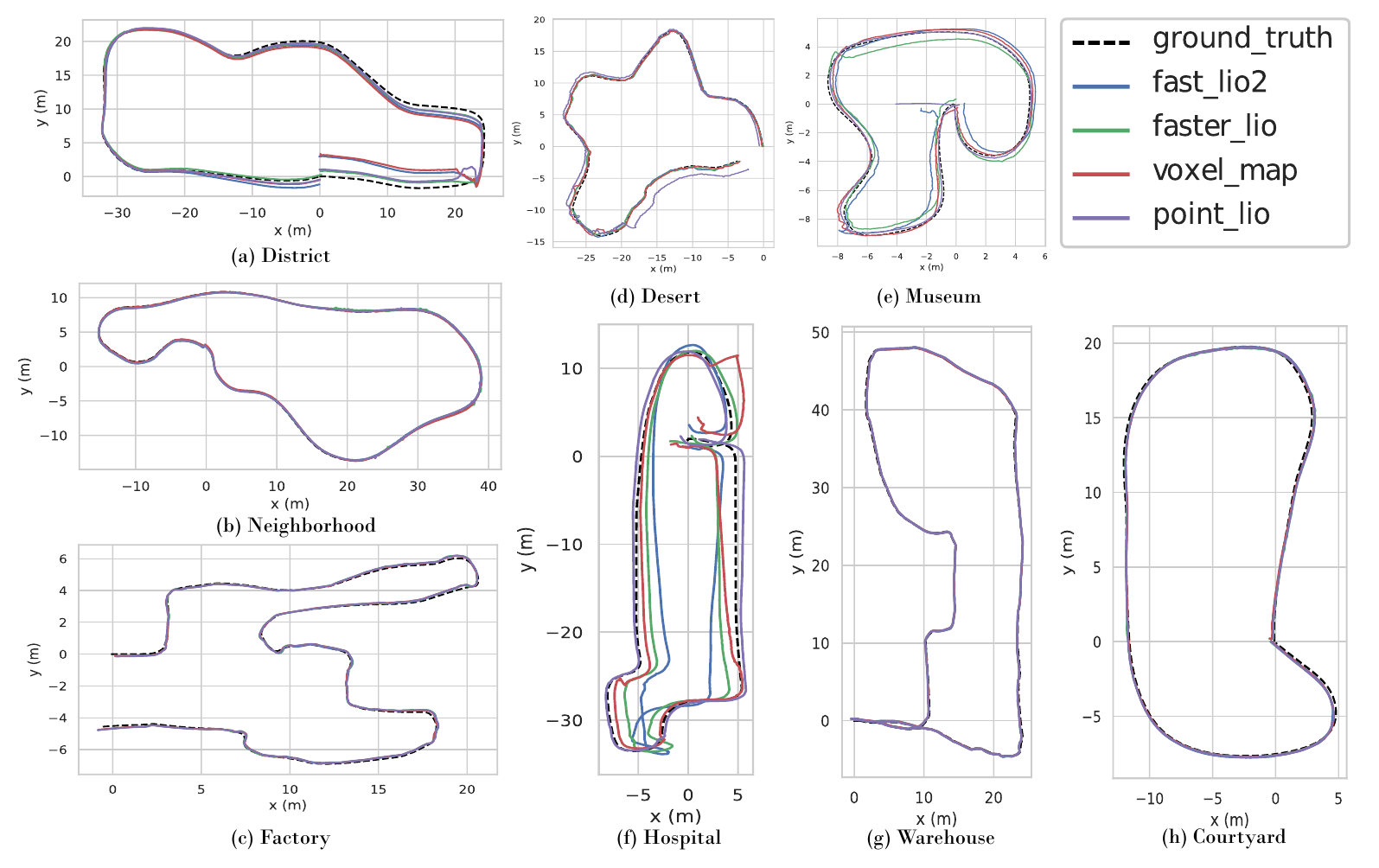}
\caption{(a)–(h) respectively show the comparison between the localization results and ground truth path in District, Neighborhood, Factory, Desert, Museum, Hospital, Warehouse, and Courtyard sequence.}
\label{path}
\end{figure*}

Table \ref{solid-state-table} presents detailed information about the test paths for solid-state LIDARs and the evaluation results on Fast-LIO2, Faster-LIO, Voxel-Map, and Point-LIO. Figure \ref{path} illustrates the comparison between estimated trajectories by solid-state LIDAR SLAM algorithms and ground truth paths. Based on the results from Table \ref{solid-state-table} and Figure \ref{path}, it can be observed that, compared to mechanical LIDARs, the localization accuracy of SLAM algorithms based on solid-state LIDAR is generally lower across all paths. Particularly, drift phenomena are more prone at narrow corners, as depicted in Figure \ref{path} for the District and Hospital sequences.

\begin{table*}[htb]
\caption{Comparison of localization accuracy of solid-state LIDAR experiments (APE in meters)}
\begin{threeparttable}
\begin{tabular}{cccccccccc}
\hline
\multicolumn{6}{c|}{\textbf{Sequence Description}}                                                                                                                   & \multicolumn{4}{c}{\textbf{Absolute Pose Error (APE)}}                             \\ \hline
\multicolumn{1}{c|}{\textbf{Index}} & \textbf{Route}  & \textbf{LIDAR-Type} & \textbf{Sequence-Type} & \textbf{Control-Mode} & \multicolumn{1}{c|}{\textbf{Feature}} & \textbf{Fast-LIO2} & \textbf{Faster-LIO} & \textbf{Voxel-Map} & \textbf{Point-LIO} \\ \hline
\multicolumn{1}{c|}{\textbf{1}}     & District        & Avia                & Outdoor                & Auto                  & Dyanmic                               & 1.179851           & \textbf{0.620550}   & 1.423089           & 0.657611           \\
\multicolumn{1}{c|}{\textbf{2}}     & Neighborhood        & Hap                 & Outdoor                & Auto                  & Structural                            & 0.228922           & 0.215239            & \textbf{0.084925}  & 0.195440           \\
\multicolumn{1}{c|}{\textbf{3}}     & Desert            & Avia                & Outdoor                & Manual                & Degenerated                           & 0.278179           & \textbf{0.251939}   & 0.283014           & 0.952304           \\
\multicolumn{1}{c|}{\textbf{4}}     & Courtyard      & Horizon             & Outdoor                & Auto                  & Structural                            & 0.198771           & 0.204139            & \textbf{0.196741}  & 0.205847           \\
\multicolumn{1}{c|}{\textbf{5}}     & Museum         & Avia                & Indoor                 & Auto                  & Structural                            & 0.739714           & 0.548547            & 0.526399           & \textbf{0.290007}  \\
\multicolumn{1}{c|}{\textbf{6}}     & Factory         & Hap                 & Indoor                 & Manual                & Structural                            & 0.238224           & 0.235053            & \textbf{0.223123}  & 0.246575           \\
\multicolumn{1}{c|}{\textbf{7}}     & Hospital & hap                 & Indoor                 & Auto                  & Structural                            & 2.717883           & 1.366048            & 1.524309           & \textbf{0.636238}  \\
\multicolumn{1}{c|}{\textbf{8}}     & Corridor & Horizon             & Indoor                 & Auto                  & Degenerated                           & 11.903684           & 3.368509   & \textbf{0.228866}           & 1.954102          \\
\multicolumn{1}{c|}{\textbf{9}}     & Warehouse       & hap                 & Indoor                 & Manual                & Dyanmic                               & 0.268302           & 0.247004            & \textbf{0.235255}  & 0.251961           \\ \hline
\end{tabular}
\label{solid-state-table}
\begin{tablenotes}
        \footnotesize
        \item The bold values represent the algorithms with lowest positioning error in the current sequence.
      \end{tablenotes}
\end{threeparttable}
\end{table*}

In narrow corners, insufficient environmental feature information, compounded by the limited FoV of solid-state LIDARs (as shown in Table 3, where the horizontal field of view of most solid-state LIDARs is significantly smaller than that of mechanical LIDARs, 360 degrees), limits the environmental information captured by the LIDAR. Insufficient observational information leads to registration deviations, exacerbating cumulative errors and thus affecting the positioning performance of SLAM algorithms in such scenarios. Additionally, in unstructured environments (e.g., Desert), the inherent lack of structural features further degrades localization performance. The limited Fov of solid-state LIDAR exacerbates the drift of the algorithm in these types of environments.

\begin{figure}
\centering
\includegraphics[width=8cm]{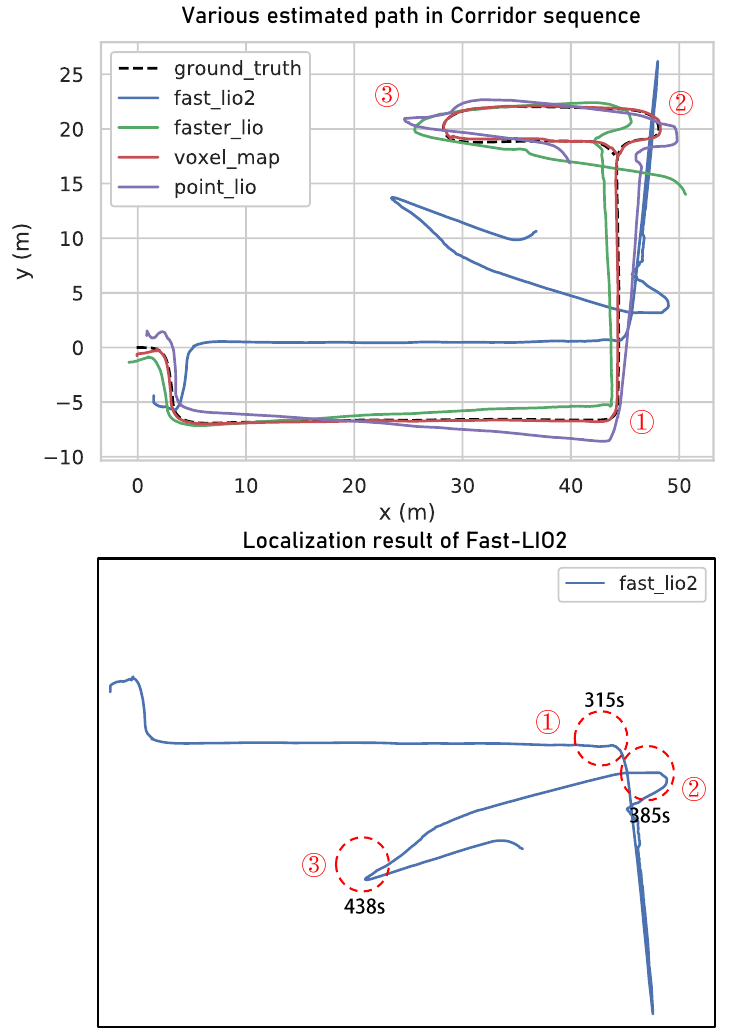}
\caption{Estimated path of various SLAM algorithms and detailed localization trajectory of Fast-LIO2 in corridor sequence.}
\label{Fast-LIO2-path}
\end{figure}

It is noteworthy that in the Corridor sequence, which depicts a degraded long corridor environment, the Fast-LIO2 algorithm exhibits significant drift. Figure \ref{Fast-LIO2-path} shows the localization trajectories of various SLAM algorithms in Corridor sequence, along with the detailed trajectory output by Fast-LIO2. From the upper part of Figure \ref{Fast-LIO2-path}, it can be observed that Fast-LIO2 experiences significant drift at narrow corners 1, 2, and 3 in the Corridor sequence, leading to a sharp decline in algorithm performance. By analyzing the execution process, it is determined that the drifts at 1, 2, and 3 occur at approximately 315s, 385s, and 438s, respectively, as shown in the lower part of Figure \ref{Fast-LIO2-path}.

\begin{figure}
\centering
\includegraphics[width=9cm]{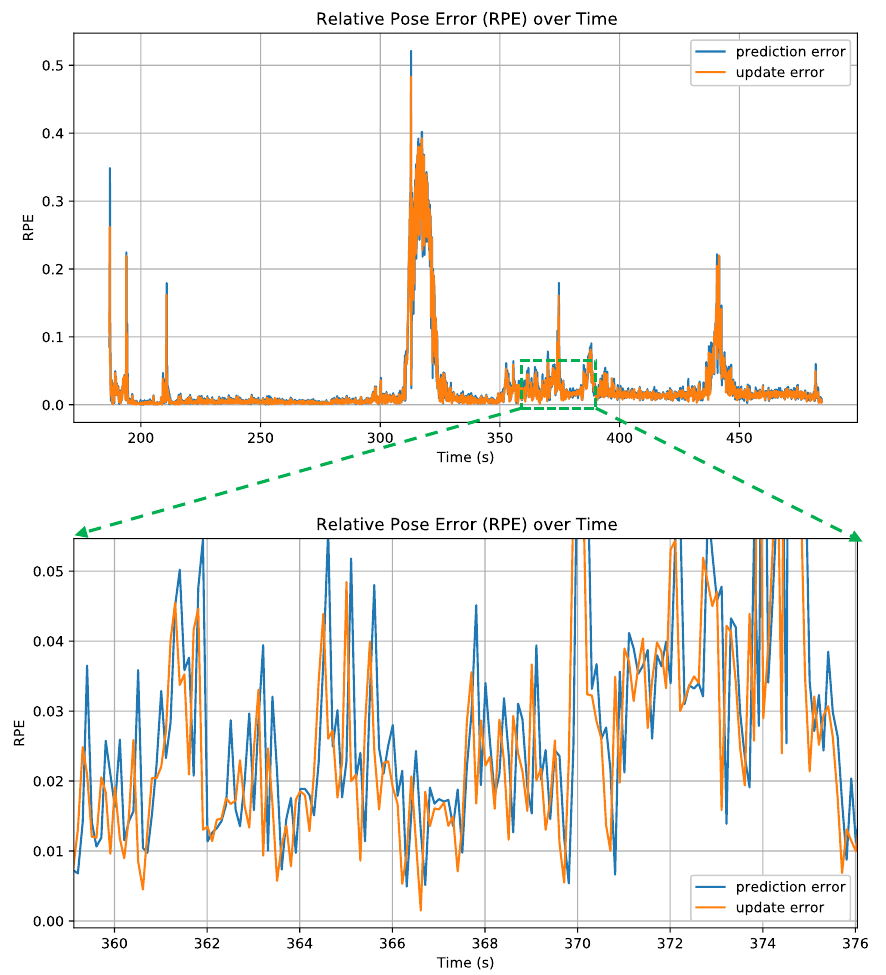}
\caption{ Comparison of RPE errors between IMU predictions and LIDAR updates relative to the ground truth.}
\label{pre_update}
\end{figure}

To determine whether the errors at these corners are caused by IMU pose prediction or LIDAR point cloud registration, we plot the relative positioning error (RPE) of the IMU prior predictions and LIDAR posterior updates relative to the ground truth, as shown in Figure \ref{pre_update}. From Figure \ref{pre_update}, it is clear that the specific times of significant system drift correspond with the results in Figure \ref{Fast-LIO2-path}. Additionally, in the detailed magnification of Figure \ref{pre_update}, the error in the updated trajectory always appears before the prediction error, indicating that the drift in the system pose at these points is initially caused by point cloud registration. This subsequent predictions based on the updated poses leads to an exacerbation of cumulative error.

\begin{figure}
\centering
\includegraphics[width=9cm]{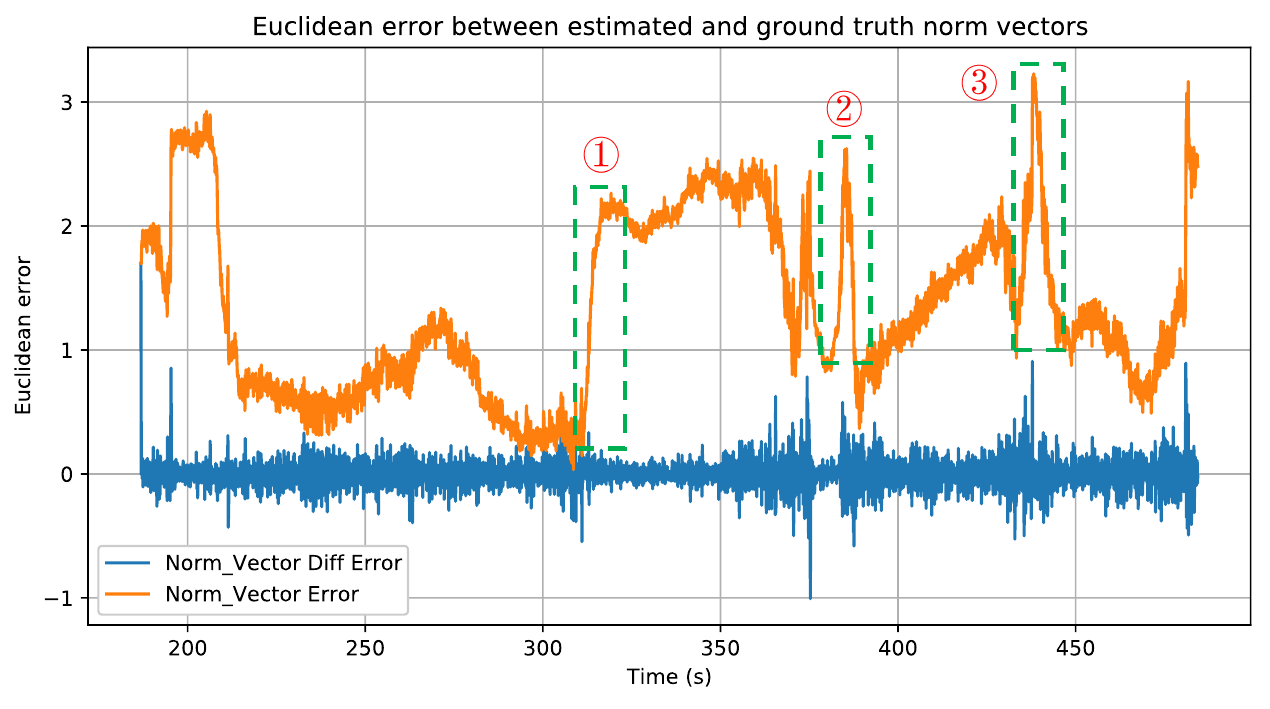}
\caption{Error Variation between the LIDAR point nearest neighbor plane normals estimated by Fast-LIO2 and the ground truth at each frame.}
\label{norm_vector}
\end{figure}

Additionally, to further verify and illustrate the causes of LIDAR registration errors and their impact on positioning results, we plot the variation of the total error between the LIDAR point nearest neighbor plane normals estimated by Fast-LIO2 and the ground truth at each frame, as shown in Figure \ref{norm_vector}. The ground truth normals are obtained by fitting the simulated error-free point clouds. As shown in Figure \ref{norm_vector}, there are significant spikes in the plane fitting errors at 315s, 385s, and 438s. This indicates that at these points, the plane fitting errors are substantial, leading to distortions in the LIDAR point cloud registration. Consequently, this causes drift in the algorithm at these locations, affecting the overall positioning performance of the algorithm.

\section{Extended research}

In this paper, we have elaborated on a multi-scenario adaptable intelligent robot simulation platform based on LIDAR-inertial fusion. This platform integrates various sensors and simulation environments, providing a versatile tool for research on robotic systems and algorithms. The platform demonstrates significant scalability, and through continuous technological upgrades and feature expansions in the future, it will further enhance its performance and application scope. It supports a wide range of needs from basic research to complex applications, laying a solid foundation for future developments. The scalability is mainly reflected in the following four aspects:

\subsection{Sensor Expansion}

    The current platform integrates various type of LIDAR and IMU sensors. In the future, this platform can add visual sensors such as cameras and infrared sensors to enhance system performance and adaptability in complex scenarios (e.g., degraded, low-light, or no-light environments)\cite{BVT-SLAM}, providing support for tasks like autonomous driving and search and rescue missions. Additionally, integrating GPS or other global positioning devices can assess the impact of inherent sensor errors on algorithm evaluation. Furthermore, considering the addition of underwater sensors such as sonar, DVL (Doppler Velocity Log) and other underwater sensors in simulation platform to support marine exploration and underwater infrastructure development\cite{holoocean,ocean_LIDAR}.

\subsection{Environment Expansion}

    The current platform offers a collection of simulated environments, which can be expanded in the future to enhance diversity and realism. The platform supports custom design and customization to generate a wider range of virtual scenarios, including underwater and marine environments, harsh weather conditions like rain and snow, and even simulations of Martian or space environments. Additionally, leveraging digital twin technology\cite{twin} to generate highly realistic virtual environments based on real-world scenes allows for the replication of dynamic changes observed in the real world. This provides a more solid foundation for the practical deployment of robot technologies.

\subsection{Carrier Morphology Expansion}

    Currently, the platform provides models of wheeled robots.In the future, this platform can expand support to various forms such as autonomous vehicles and unmanned aerial vehicles (UAVs). This will enrich the platform's applications in urban and highway environments as well as low-altitude scenarios. Additionally, the platform will include specialized robot models such as robotic dogs and legged robots to adapt to complex terrains and diverse environments, including mountains, ruins, and dense forests, for exploration and operations.

\subsection{Platform Functionality Expansion}

    Currently, the platform's functionalities are primarily focused on evaluating localization accuracy. In the future, this platform can add functionality for evaluating mapping accuracy as well. A significant advantage of simulation environments lies in providing accurate ground truth maps, which serve as an ideal benchmark for evaluating mapping algorithms. In the future, detailed testing and evaluation of mapping effectiveness for algorithms including SLAM, Nerf(Neural Radiance Fields)\cite{nerf}, and 3DGS(3D Gaussian Splatting)\cite{3dgs} can be conducted through integration or development of mapping evaluation methods.

\section{CONCLUSIONS}

This paper presents a versatile multi-scenario adaptable intelligent robot simulation platform based on LIDAR-Inertial fusion. The platform includes an intelligent robot model equipped with a sensor suite, capable of freely moving in the simulation environment through manual control or autonomous tracking. The platform supports the convenient import and custom creation of various simulation environments, including structured, unstructured, dynamic, and degraded environments. Additionally, the platform provides ground truth information with absolute accuracy, facilitating detailed analysis and evaluation of various SLAM algorithms. Experiments conducted with mechanical and solid-state LIDARs demonstrate the extensive adaptability and practicality of our platform. Furthermore, the platform exhibits strong extensibility in sensor simulation, environment creation, and map evaluation. 

In the future, we plan to develop a GPU-based solid-state LIDAR simulation plugin to enhance the simulation efficiency of the platform in large-scale virtual scenarios.

\addtolength{\textheight}{-8cm}   










\bibliographystyle{unsrt}
\bibliography{ref.bib}

\end{document}